\documentclass[letterpaper, 10 pt, conference]{IEEEtran}
\IEEEoverridecommandlockouts
\usepackage{balance}
\usepackage{cite}
\usepackage{amsmath,amssymb,amsfonts,bbm,mathtools}
\usepackage{hyperref}
\usepackage[ruled]{algorithm2e}
\usepackage{graphicx,subcaption}
\usepackage{textcomp}
\usepackage{xcolor}
\usepackage{comment}
\usepackage{multirow}
\def\BibTeX{{\rm B\kern-.05em{\sc i\kern-.025em b}\kern-.08em
    T\kern-.1667em\lower.7ex\hbox{E}\kern-.125emX}}
\begin{document}

\title{\LARGE \bf Bimanual Regrasping for Suture Needles using Reinforcement Learning for Rapid Motion Planning}

\author{
Zih-Yun Chiu$^1$, Florian Richter$^1$ \IEEEmembership{Student Member, IEEE}, \\Emily K. Funk$^2$, Ryan K. Orosco$^2$ \IEEEmembership{Member, IEEE}, and Michael C. Yip$^1$ \IEEEmembership{Member, IEEE}
\thanks{This research was supported by the Telemedicine and Advanced Technology Research Center (TATRC) T.R.O.N. program.}%
\thanks{$^1$Zih-Yun Chiu, Florian Richter, and Michael C. Yip are with the Department of Electrical and Computer Engineering, University of California San Diego, La Jolla, CA 92093 USA. {\tt\small \{zchiu, frichter, yip\}@ucsd.edu}}% 
\thanks{$^2$Emily K. Funk and Ryan K. Orosco are with the Department of Surgery - Division of Head and Neck Surgery, University of California San Diego, La Jolla, CA 92093 USA. {\tt\small \{ekfunk, rorosco\}@ucsd.edu}}%
}

\maketitle
\begin{abstract}
Regrasping a suture needle is an important yet time-consuming process in suturing. 
To bring efficiency into regrasping, prior work either designs a task-specific mechanism or guides the gripper toward some specific pick-up point for proper grasping of a needle. 
Yet, these methods are usually not deployable when the working space is changed. 
Therefore, in this work, we present rapid trajectory generation for bimanual needle regrasping via reinforcement learning (RL). 
Demonstrations from a sampling-based motion planning algorithm is incorporated to speed up the learning. 
In addition, we propose the ego-centric state and action spaces for this bimanual planning problem, where the reference frames are on the end-effectors instead of some fixed frame. 
Thus, the learned policy can be directly applied to any feasible robot configuration. 
Our experiments in simulation show that the success rate of a single pass is 97\%, and the planning time is 0.0212s on average, which outperforms other widely used motion planning algorithms. 
For the real-world experiments, the success rate is 73.3\% if the needle pose is reconstructed from an RGB image, with a planning time of 0.0846s and a run time of 5.1454s. 
If the needle pose is known beforehand, the success rate becomes 90.5\%, with a planning time of 0.0807s and a run time of 2.8801s. 
\end{abstract}

\section{Introduction}

\begin{figure}[t!]
    \centering
    \includegraphics[width=\linewidth]{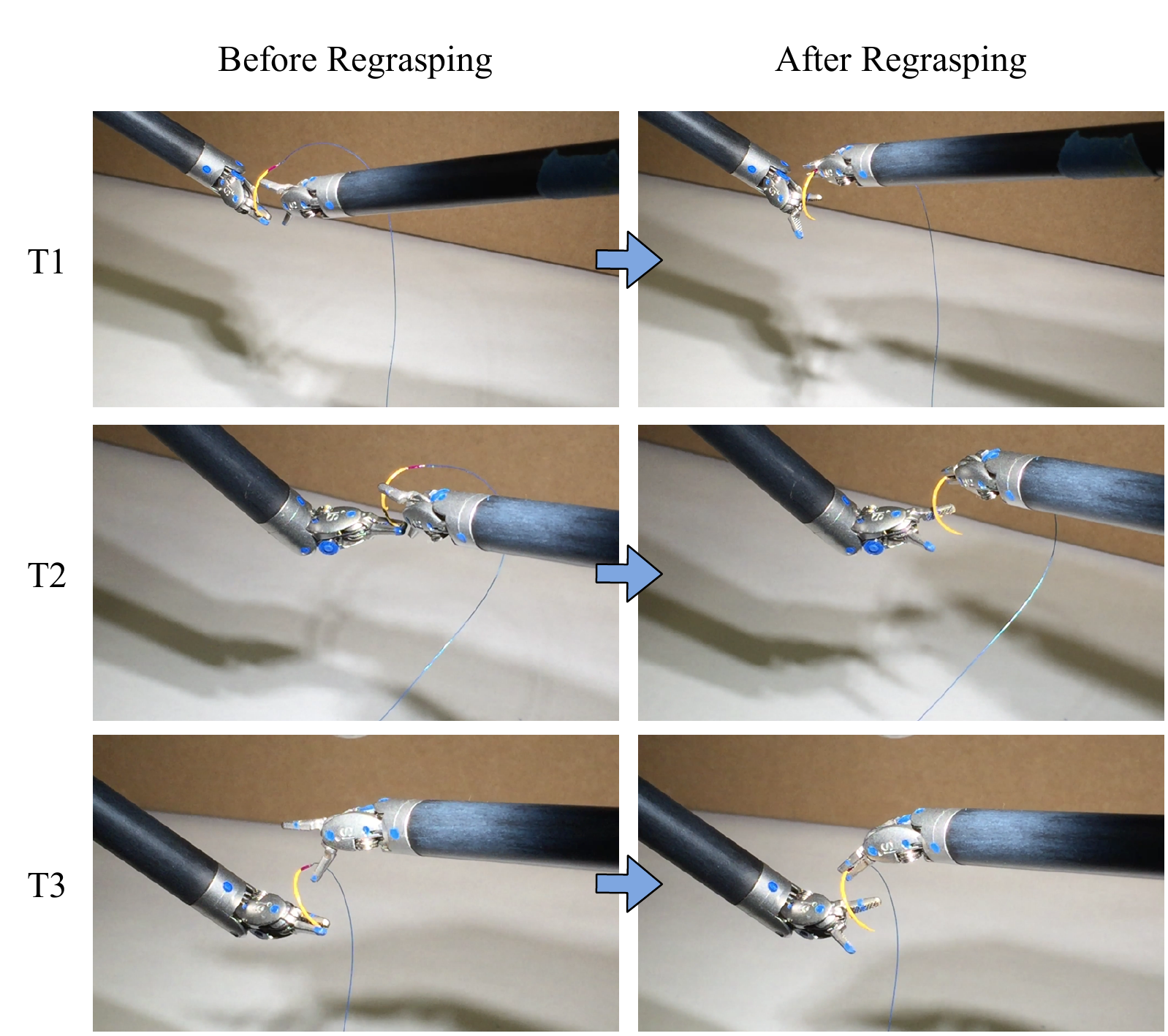}
    \caption{Three trials (T1, T2, T3) of successful needle regrasping. The pictures on the right show a da Vinci surgical robot arm successfully regrasps the suture needle at the pose where a surgeon often holds before suturing.}
    \label{fig:cover_figure}
\end{figure}

On the journey towards complete autonomy in surgical robotics \cite{yipDasJournal}, automation of suturing has engrossed the community. 
There is a good reason for this, as manual suturing has been cited as being time consuming~\cite{hubens2003performance}, and suturing on Robot Assisted Minimally Invasive Surgeries (RAMIS) can be tedious and challenging~\cite{corcione2005advantages}. 
Hence, automation of suturing becomes an attractive goal for improving a surgeon's quality of life during RAMIS. 
Emergence of surgical robotics simulators \cite{richter2019open} and open-sourced robotic platforms such as the da Vinci Research Kit (dVRK) \cite{kazanzides2014open} provide the ability for rapid development towards realizing this goal. 

% How is it done today, and what are the limits of current practice
Suturing requires a wide range of considerations for effective automation including needle manipulation \cite{sundaresan2019automated}, knot tying \cite{mayer2008system}, identification of entry and exit points for the suture throw \cite{liu2016needle}, and interfacing the automation with the surgeon for effective deployment \cite{watanabe2017single}.
In this work, we investigate the specific sub-challenge of needle manipulation from the perspective of bimanual regrasping, which is to regrasp the needle at a proper pose using dual robot arms.
Fontanelli et al. cite that 74\% of suture throws from the in-vivo JIGSAW dataset \cite{gao2014jhu} required regrasping which took an expert on average 7.9s to complete \cite{fontanelli2018new}.
Previous work has also shown that proper grasping can enhance the effectiveness of suture throws  \cite{liu2016needle, liu2015optimal}.
In fact, many planning techniques for suture needle throwing have even used task-specific gripper mechanisms to ensure proper grasping of a needle \cite{jackson2013needle, leonard2014smart, sen2016automating, pedram2017autonomous}.
While these mechanisms are effective in highlighting suture throwing capabilities, they are too task-specific to be deployed on a surgical robot.

On the other hand, previous work has approached the proper needle grasping problem by visual servoing~\cite{d2018automated} or learning from demonstrations~\cite{varier2020collaborative}. 
However, these works might not avoid needle regrasping after picking it up if the needle is displaced or reoriented due to collision. 
Also, these methods are designed under some specific robot configuration and might require new calibration or training in another working space. 
Therefore, regrasping suture needles to the optimal location for throwing on standard RAMIS tooling is still of great importance.% towards eventually aiding surgeons when suturing.

Standard practice for surgeons is to regrasp the suture needle roughly one third needle length from the suture thread and over $30^\circ$ off the needle as shown in Fig. \ref{fig:cover_figure}.
This gives an end-goal for the robot to achieve that can be given to a motion planner as a terminal state in order to find a collision free trajectory for regrasping.
However, our experiments show that every available state-of-the-art motion planner was too slow for this task and often times-out. 
This problem is not unexpected; planning times typically can take up to 3 minutes for constrained environments and neural planning methods are required to breach this limitation \cite{qureshi2020motion}.
These methods however are not directly developed for dual arm robots which are typical for RAMIS.
In particular, the bases of the robotic arms are regularly adjusted between procedures so the learning based motion planner needs to generalize to all potential workspaces of the surgical robot.

% What is new in your approach and why do you think it will be successful?
To this end, we present a novel method for bimanual regrasping for suture needles. We use an RL approach integrated with rapid replanning to demonstrate real-time performance and success in regrasping. We structure the problem to be ego-centric to the needle and the end-effectors, making it far more generalizable than a typical configuration-space approach. 

To plan for the needle passing task, the states and actions are related to the pose of the two end-effectors. 
The reference frames for these poses are not set to be the world frame or the base frame of an arm. 
Instead, the reference frames can be the frames of the end-effectors, which \textit{change as they move}. 
This setting is referred as \textit{the ego-centric setting}, since the information that the end-effectors get at each time step is centered on themselves.

There are several benefits of using the ego-centric setting. 
First, this setting allows the policy to successfully work in the global space without requiring the agent to explore the wide range of space. 
Instead, it only needs to observe the states relative to the goal, which are included in a local space. 
Hence, no matter where the agent and the goal are in the world frame, as long as their relation has been learned by the policy, the agent is able to reach the goal. 
Also, since the policy does not consider the configuration of a robot, it can be directly used for other robot arms. 
These benefits suggest that the ego-centric setting is very data efficient for learning a policy that can work in many different scenarios.

To sum up, we specifically present the following contributions in this work:
\begin{enumerate}
    \item fast trajectory generation for bimanual needle regrasping for suture throwing via RL,
    \item RL training strategy which incorporates intermittent, targeted exploration to guide the policy while still allowing for it to generalize, and
    \item ego-centric parameterization of the needle and surgical tools to generalize the trajectory for non-specific end-effector nor robotic base positioning.
\end{enumerate}
%A simulated environment for the bimanual needle regrasping task is developed and open sourced on dVRL\footnote[2]{dVRL available at \url{https://github.com/ucsdarclab/dVRL}} \cite{richter2019open} to prototype and test trajectory generation techniques.
%The trajectories are tested on dVRK \cite{kazanzides2014open} with a 5mm radius needle using real-time surgical tool and needle tracking resulting in a ??\% success rate.
While this work is directly aimed towards reducing the time needed to suture when conducting RAMIS, we also foresee it being an important step towards making other automated suture throw techniques deployable by relieving them of task-specific suture needle grasping mechanisms.

\section{Reinforcement Learning Background}
The suture needle passing task in this work is formulated as an RL problem and is modelled as a Markov Decision Process (MDP). 
The goal is to find a deterministic policy $\pi$ that solves this MDP. 
The following notations and definitions are used for the RL formulation. 
Let $\mathcal{M} = (\mathbb{S}, \mathbb{A}, P, r, \rho_0, \gamma, T)$ be the tuple for the discrete-time finite-horizon MDP, where $\mathbb{S}$ is the state space, $\mathbb{A}$ is the action space, $P$ is the transition probability, $r$ is the reward function, $\rho_0$ is the initial state distribution, $\gamma$ is the discount factor, and $T$ is the task horizon. 
A policy $\pi$ is learned to maximize the expected return $\mathbb{E}_{\mathbf{s}_0\sim\rho_0, P}[\sum_{t=0}^T \gamma^t r(\mathbf{s}_t, \mathbf{a}_t)]$, where $\mathbf{s}_t\in\mathbb{S}$ is the state at time step $t$, $\mathbf{a}_t = \pi(\mathbf{s}_t) \in \mathbb{A}$ is the action at $t$, and $\mathbf{s}_{t+1}\sim P(\mathbf{s}_{t+1}| \mathbf{s}_t, \mathbf{a}_t)$. 

\section{Methods}
In order to develop and test suture needle passing, a simulated environment is developed in V-REP based on Fontanelli's et al. previous work \cite{vrep_simulator}.
\begin{comment}
The simulated scene requires an initialized grasp and a goal grasping point which is first described in the coming subsection.
From there, the ego-centric setting used for the states and actions of planning needle passing is proposed. 
Then how expert demonstrations are generated and an RL policy is trained for passing a needle in the simulated scene is explained.
\end{comment}

\subsection{Initialization and Goal Generation for Needle Grasping}
\label{sec:initialization}
At the beginning of conducting needle passing, the needle can be grasped at a variety of points and directions.
Therefore, the RL environment will be initialized with a random sample of an initial grasping point and direction such that the trained policy for motion planning will generalize well.
In addition, the grasping point must be well defined.
To model this mathematically, the following two coordinate frames are built on the needle. 

\begin{figure}[t]
    \centering
    \vspace{2mm}
    \begin{subfigure}{0.225\textwidth}
        \centering
        \hspace*{3.5cm}
        \includegraphics[width=\textwidth]{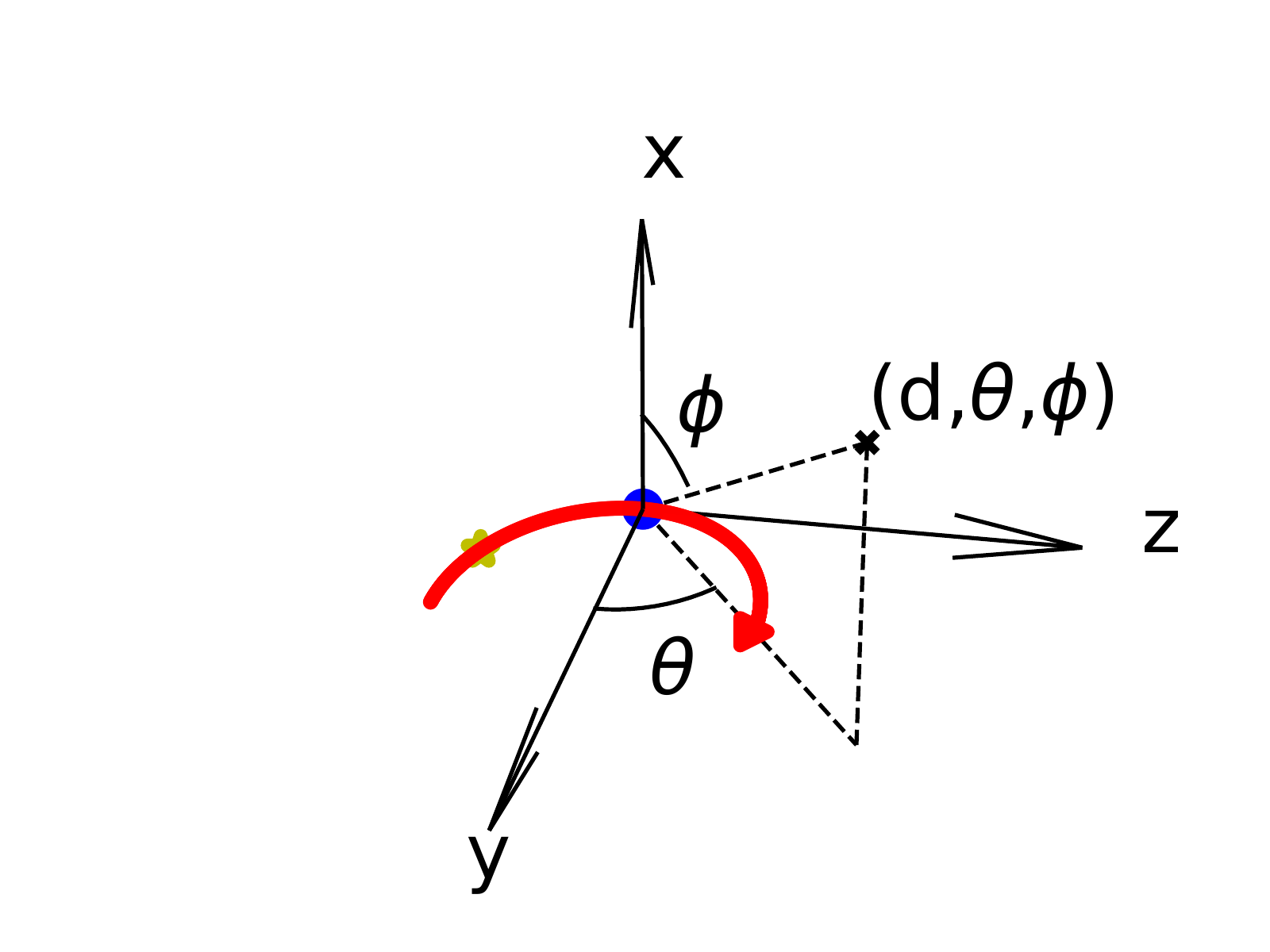}
        \caption{Needle frame}
        \label{fig:needle_frame}
    \end{subfigure}%
    \begin{subfigure}{0.225\textwidth}
        \centering
        \hspace*{-8.5cm}
        \includegraphics[width=\textwidth]{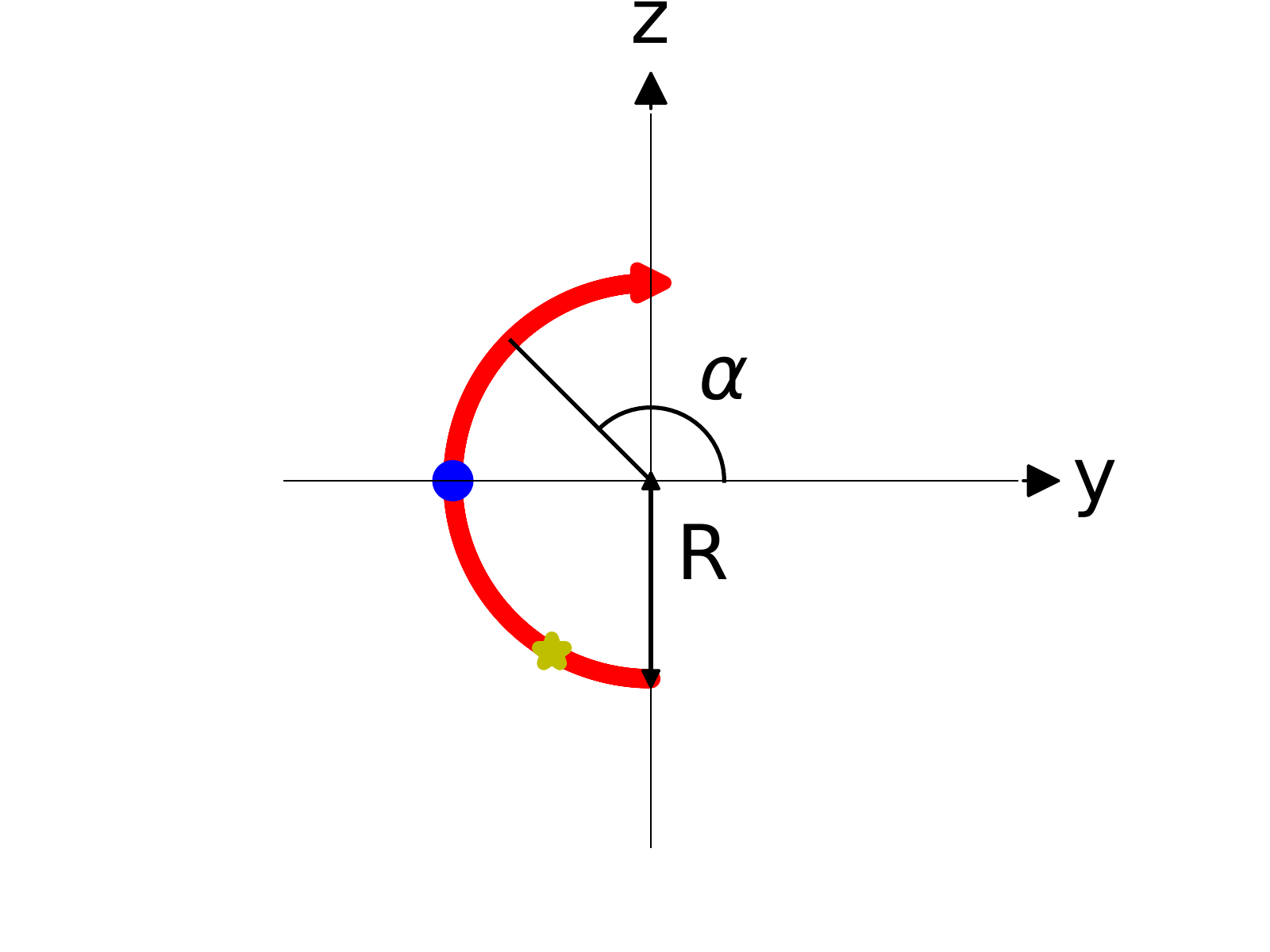}
        \caption{Grasping point frame}
        \label{fig:grasping_point_frame}
    \end{subfigure}
    \caption{Coordinate systems on the needle (red semi-circle: the suture needle, red arrow: the tip of the suture needle, blue dot: an example of a grasping point, yellow dot: an example of a goal grasping point)}
    \label{fig:frames}
\end{figure}

\subsubsection{Needle frame}
A Cartesian coordinate system is defined on the suture needle as shown in Fig.  \ref{fig:needle_frame}.
This frame is referred as \textit{the needle frame}.
The most common suture needle used in RAMIS are of a semicircle shape, hence these will be considered in this work.
The equation for the semicircle suture needle in the needle frame is defined as: 
\begin{equation}
    \label{equ:gp_equ}
    \left\{
    \begin{array}{l}
        x_n = 0 \\
        y_n = R\cos\alpha \\
        z_n = R\sin\alpha.
    \end{array}
    \right .
\end{equation}
where $R$ is the radius of the needle and $\alpha\in\left[\frac{\pi}{2},\frac{3\pi}{2}\right]$ is the angle on the needle. 
By randomly sampling an $\alpha$, a grasping point on the suture needle can be sampled.
The pose of this grasping point in the needle frame is denoted as $\left( \mathbf{p}^n_g, \mathbf{q}^n_g \right)$, where $\mathbf{p}^n_g = \left[x_n\ y_n\ z_n\right]^\top$ and is calculated by equation (\ref{equ:gp_equ}). Here, $\mathbf{q}^n_g = \left[1\ 0\ 0\ 0\right]^\top$ is the identity rotation represented in the quaternion form.
Likewise, the goal grasping point can be defined by setting $\alpha$.

\subsubsection{Grasping point frame}

In order to sample a grasping direction pointing to a grasping point for initialization, a spherical coordinate system is defined with the origin at the grasping point as shown in Fig. \ref{fig:grasping_point_frame}.
This frame is referred as \textit{the grasping point frame}.
A point using Cartesian representation in this frame can be calculated as
\begin{equation}
    \label{equ:gd_equ}
    \left\{
    \begin{array}{l}
        x_g = d\cos\phi \\
        y_g = d\sin\phi\cos\theta \\
        z_g = d\sin\phi\sin\theta.
    \end{array}
    \right .
\end{equation}
where $(d, \theta, \phi)$ are the radius, azimuth, and inclination respectively from the grasping point frame.
This parameterization is beneficial for setting grasping direction of a robotic gripper.
$d$ defines the depth of the grasp with regards to the gripper while $\theta$ and $\phi$ give the grasping angle relative to the needle.
Therefore, by randomly sampling a $(d, \theta, \phi)$, the target position of the end-effector will be set to $\mathbf{p}^g_e = [x_g\ y_g\ z_g]^\top$, and its orientation will become $\mathbf{q}^g_e$ such that the gripper points from $[x_g\ y_g\ z_g]^\top$ to the origin in the grasping point frame.
%After $\left( \mathbf{p}^g_e, \mathbf{q}^g_e \right)$ is sampled, it can be transformed to the needle frame by 
$\left( \mathbf{p}^g_e, \mathbf{q}^g_e \right)$ can then be transformed to the needle frame by 
\begin{equation}
    \label{equ:gp_gd_needle_frame}
    \mathbf{H}^n_e \left( \mathbf{p}^n_e, \mathbf{q}^n_e \right) = 
    \mathbf{H}^n_g \left( \mathbf{p}^n_g, \mathbf{q}^n_g \right) 
    \mathbf{H}^g_e \left( \mathbf{p}^g_e, \mathbf{q}^g_e \right). 
\end{equation}
where $\mathbf{H}(\cdot, \cdot) \in SE(3)$ is the homogeneous representation of a pose.
Then the end-effector can be set to reach $\left( \mathbf{p}^n_e, \mathbf{q}^n_e \right)$, hence grasping the needle, through inverse kinematics. 
The goal grasping direction can be set in a similar fashion.
In the following sections, the end-effector that is initialized to hold a needle is referred as the grasping end-effector, and the one that approaches the goal is referred as the regrasping end-effector.

\subsection{Ego-Centric State and Action Space}

\begin{figure}[t!]
\centering
    \vspace{1.9mm}
    \includegraphics[height=1.85cm]{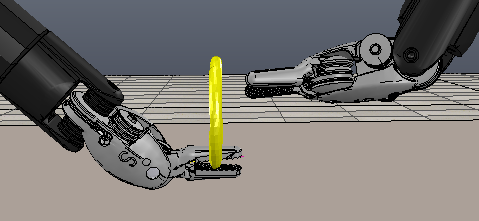}
    \includegraphics[height=1.85cm]{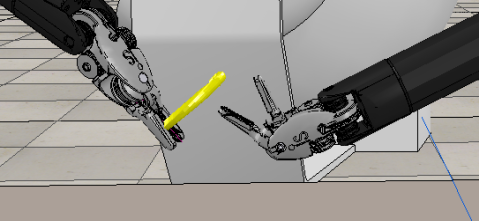}
    \caption{Two different configurations of the robot arm that have the same state. Since a state is defined with the ego-centric setting, no matter how the robot configuration changes, as long as the relative poses of the two end-effectors and the needle remain the same, the state will not be changed.}
    \label{fig:ego_centric_configs_vrep}
\end{figure}

With the ego-centric setting, the states in $\mathbb{S}$ and actions in $\mathbb{A}$ for the needle passing task are defined as follows. 
A state includes 
\begin{enumerate}
    \item the position and quaternion of the needle measured in the grasping end-effector frame $\mathbf{p}^{e_g}_n$, $\mathbf{q}^{e_g}_n$, and 
    \item the position and quaternion of the regrasping end-effector measured in the grasping end-effector frame $\mathbf{p}^{e_g}_{e_r}$, $\mathbf{q}^{e_g}_{e_r}$, and 
    \item the position and quaternion of the needle measured in the regrasping end-effector frame $\mathbf{p}^{e_r}_n$, $\mathbf{q}^{e_r}_n$. 
\end{enumerate}
After initializing the environment, $( \mathbf{p}^{e_g}_n, \mathbf{q}^{e_g}_n )$ is available by calculating the inverse of equation (\ref{equ:gp_gd_needle_frame}). 
For simplicity, in this work it is assumed that only the regrasping arm moves during planning. 
Yet, the ego-centric setting can be directly applied to two moving arms. 
Moreover, since a state has nothing to do with the joint angles or the base frame of the robot, two very different configurations of a robot arm can have the same state as shown in Fig. \ref{fig:ego_centric_configs_vrep}. 

An action is defined as the variation of the position and quaternion of the regrasping end-effector measured in the grasping end-effector frame $\Delta\mathbf{p}^{e_g}_{e_r}$, $\Delta\mathbf{q}^{e_g}_{e_r}$. 
Given that the regrasping end-effector moves from $(\mathbf{p}^{e_g}_{e_r,t-1}, \mathbf{q}^{e_g}_{e_r,t-1})$ to $(\mathbf{p}^{e_g}_{e_r,t}, \mathbf{q}^{e_g}_{e_r,t})$ at time step t, the action would be
\begin{align}
    \label{equ:position_action}
    \Delta\mathbf{p}^{e_g}_{e_r} & = \mathbf{p}^{e_g}_{e_r,t} - \mathbf{p}^{e_g}_{e_r,t-1} \\
    \Delta\mathbf{q}^{e_g}_{e_r} & = \mathbf{q}^{e_g}_{e_r,t}\  \left( {\mathbf{q}^{e_g}_{e_r,t-1}} \right) ^{-1} 
    \label{equ:quaternion_action}
\end{align}
which would eventually be used as the control commands to plan with in this ego-centric setting. 

\subsection{Learning a Policy by Deep Deterministic Policy Gradients with Demonstrations}
\label{sec:ddpg_w_demo}

%\textcolor{red}{Add high level explanation of how we are using RL to solve the motion planning task. The specifics here would be the reward function. The deeper explanation would be how the reward function "solves" motion planning with RL}
To solve the suture needle passing task, we train a policy by Deep Deterministic Policy Gradients (DDPG) with behavior cloning (BC)~\cite{nair2018overcoming}.
DDPG is one of the most widely used RL algorithms for continuous control, and its combination with BC helps guide the exploration of an agent. 
Since the process of passing a needle is very sophisticated, especially when the two end-effectors are close to each other, it is reasonable to incorporate demonstrations into learning to motivate precise motions and speed up training. 

The states and actions of this task are defined by the ego-centric setting as mentioned in the previous section. 
Also, the rewards are defined as follows~\cite{jurgenson2019harnessing}:
\begin{equation}
    r = \left\{
    \begin{array}{cl}
        -1, & \text{if collision happens} \\
        1, & \text{if equation (\ref{equ:pos_diff}) and (\ref{equ:quat_diff}) are satisfied}  \\
        \beta, & \text{otherwise}
    \end{array}.
    \right .
\end{equation}
\begin{align}
    \label{equ:pos_diff}
    & \lVert \mathbf{p}^{e_r}_{n,*} - \mathbf{p}^{e_r}_{n,t} \rVert_2 \leq \varepsilon_p \\
    \label{equ:quat_diff}
    & \lVert \text{AxisAngle}(\mathbf{q}^{e_r}_{n,*}\ (\mathbf{q}^{e_r}_{n,t})^{-1}) \rVert_2 \leq \varepsilon_q
\end{align}
Equation (\ref{equ:pos_diff}) and (\ref{equ:quat_diff}) describe the distance between the current pose $(\mathbf{p}^{e_r}_{n,t}, \mathbf{q}^{e_r}_{n,t})$ and the goal grasping pose $(\mathbf{p}^{e_r}_{n,*}, \mathbf{q}^{e_r}_{n,*})$, where $\varepsilon_p \geq 0$ and $\varepsilon_q \geq 0$ are the maximum values of this distance. 
$\text{AxisAngle}(\cdot)$ transforms a quaternion to its axis-angle representation. 
$\beta < 0$ is a value that can be tuned. 
This value should be small enough to prevent the agent from not approaching the needle, while it cannot be too small, or the agent might choose to collide with the needle instead of roaming to find a feasible path. 
Usually, the value of $\beta$ is tuned according to the reward of collision, which is -1 here, and the maximum time steps that the agent is allowed to run in the environment. 

\subsubsection{Generating expert demonstrations}

Generating expert demonstrations can be time consuming if we ask an real expert to do so. 
However, the goal position and orientation of holding a suture needle can be calculated from equation (\ref{equ:gp_gd_needle_frame}) by setting some target $\left( \mathbf{p}^n_{e_r}, \mathbf{q}^n_{e_r} \right)$. 
Therefore, the expert demonstrations can be generated automatically in the simulation environment. 

An intuitive way to do this is to apply some motion planning algorithm to a given start and goal configurations. 
Yet, in this way, the feasible space is narrower near the goal, so a gripper is more likely to collide with the needle or another gripper. 
This leads to failure for planning algorithms to find a feasible path in a given time period. 

\begin{figure}[t!]
\centering
    \vspace{2mm}
    \includegraphics[trim = 0cm 0cm 0cm 2.1cm, clip, height=3.65cm]{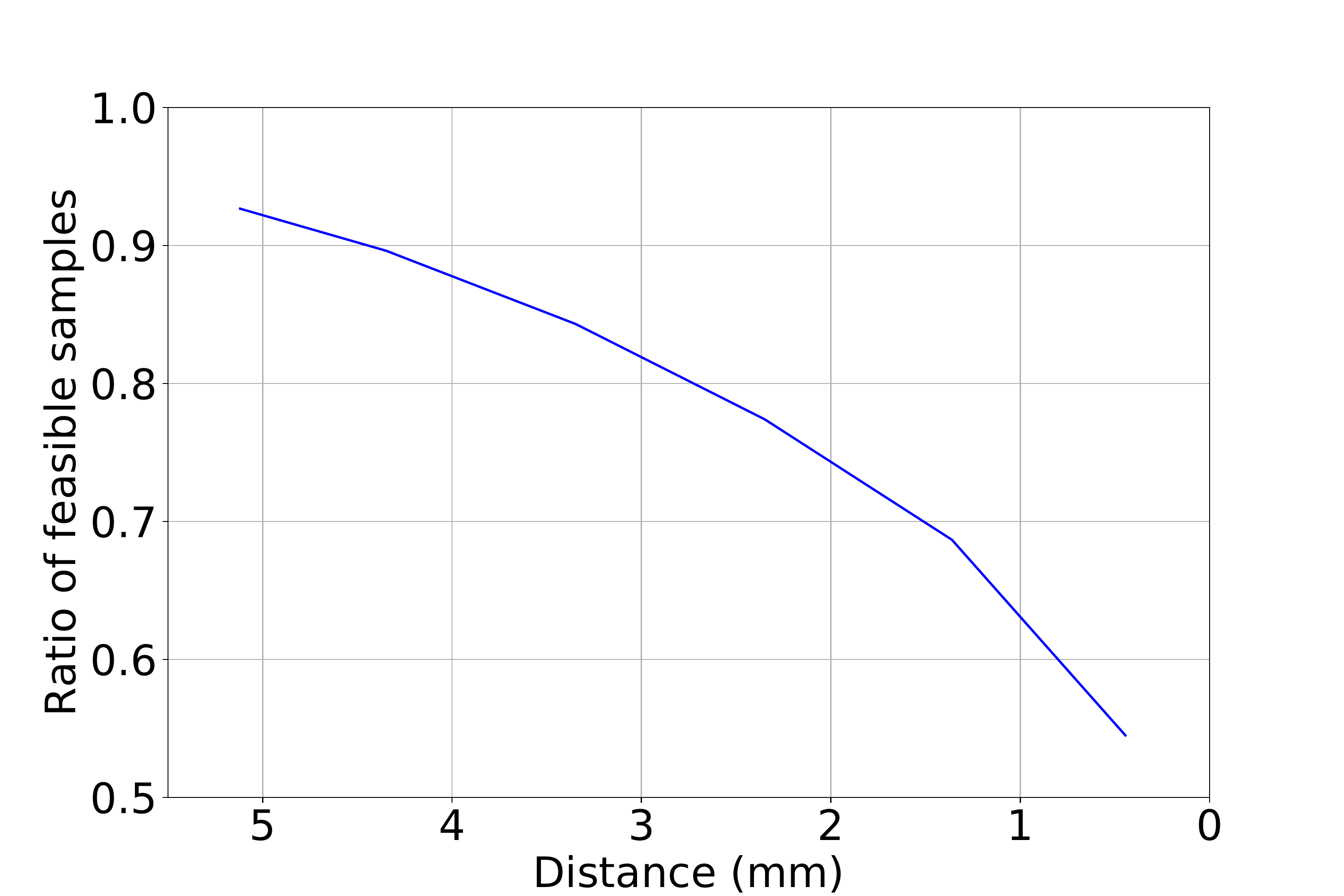}
    \includegraphics[height=3.7cm]{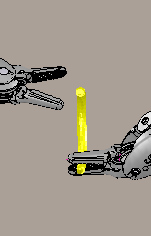}
    \caption{Feasible space and scenario when the regrasping end-effector approaches the needle. As being closer to the needle, the collision-free space decreases in a funnel like fashion as shown by the plot on the left. Therefore, forward planning is very challenging, and instead we use reverse planning.
    % This forces the initial configuration to be the toughest one, and the rest of the planning becomes easier.
    }
    \label{fig:scenario_feasible_space_approach_needle}
\end{figure}

Fig. \ref{fig:scenario_feasible_space_approach_needle} shows the scenario and probability of no collision when the regrasping gripper moves closer to the goal. 
From this figure, it can be observed that the feasible space looks like a funnel with the narrow part near the goal. 
Note that this space shrinks a lot within a small distance hence making the planning require high precision. 
This makes a motion planning algorithm more difficult to find a path without collision.
To overcome this difficulty, instead of planning from a starting point to a goal grasping point, a path can be planned in reverse, i.e., escaping a funnel.
%Since it forces the starting point to be set in the toughest situation, the rest of the planning becomes easier. 
Based on this concept, the expert demonstrations are generated by applying a motion planning algorithm from the goal grasping point to a randomly sampled starting point in the free space. 
The planned path is then reversed to be used for learning a policy. 

%TODO: add a sentence saying how the feasible space shrinks within such a short distance
% \begin{figure}[t!]
% \centering
%     \begin{subfigure}[t]{0.2\textwidth}
%     \centering
%         \includegraphics[height=3cm]{images/env/initialization_1.png}
%         \caption{Scenario}
%         \label{fig:scenario_approach_needle}
%     \end{subfigure}
%     \begin{subfigure}[t]{0.3\textwidth}
%     \centering
%         \includegraphics[height=3cm]{images/config_5001_5100.eps}
%         \caption{Feasible space}
%         \label{fig:feasible_space_approach_needle}
%     \end{subfigure}
%     \caption{Scenario and feasible space when the regrasping end-effector approaches the needle. As being closer to the needle, the collision-free space decreases in a funnel like fashion as shown by the plot on the right. Therefore, forward planning is very challenging, and instead we use reverse planning. This forces the initial configuration to be the toughest one, and the rest of the planning becomes easier.}
%     \label{fig:scenario_feasible_space_approach_needle}
% \end{figure}

\subsubsection{Applying demonstrations via an Active Learning approach}
%Given the difficulty of solving the needle passing task, an RL algorithm like DDPG + BC~\cite{nair2018overcoming} either needs a large amount of pre-collected demonstrations or extremely long training time. 
%Therefore, this algorithm is modified by collecting more expert demonstrations during training to achieve efficient learning. 

Active learning involves collecting demonstrations during training. It is used to demonstrate successful episodes only when the RL policy fails, which helps guide the exploration of the robot and significantly reduce random exploration~\cite{jurgenson2019harnessing}. 
%In their work, expert data is provided from a motion planning algorithm for failed episodes. 
%This is interpreted as target exploration, which guides the exploration of an agent. 
%Also in their work, it is suggested that with targeted exploration, the requirement of random exploration can be largely diminished. 
%By seeking demonstrations only in failed episodes, we can significantly reduce the requirement. 
%This is interpreted as target exploration.

This strategy, referred as target exploration in~\cite{jurgenson2019harnessing}, can be easily applied to our work. 
However, in our experiments it is observed that eliminating random exploration slows down the learning. 
The main reason is that the expert, which is a motion planning algorithm, is not perfect. 
It is not guaranteed that this algorithm can always provide a feasible path, and relying almost entirely on targeted exploration can instead hurt the performance. 
%Moreover, although not enough to learn a useful policy, DDPG + BC might already achieve some successes with random exploration. 
Therefore, to make an agent explore in a more effective way, random exploration is kept, and target exploration is gradually introduced to training. 
This exploration strategy is referred as \textit{mixed exploration}. 

When generating episodes with mixed exploration, if an episode fails, then with probability $\mathcal{E}_T$, a motion planning algorithm will generate a demonstration for this episode. 
%This is the process of target exploration. 
$\mathcal{E}_T$ is the probability of applying target exploration and will gradually increase from $\mathcal{E}_T^0$ to 1. 
Meanwhile, to stabilize training, the probability of random exploration $\mathcal{E}_R$ will gradually decrease from $\mathcal{E}_R^0$ to 0. 
%This mixed exploration strategy also have diminishing effect, since target exploration will be applied less frequently as the policy gradually becomes better~\cite{jurgenson2019harnessing}. 

\begin{algorithm}[t!]
  \SetAlgoLined
  \KwIn{env, number of episodes $N$, policy $\pi$, random exploration probability $\mathcal{E}_R$, target exploration probability $\mathcal{E}_T$, motion planning algorithm MP}
  $\textbf{episodes} \leftarrow [\ ]$ \\
  \For{$n \leftarrow 1$ to $N$}{
    $\textbf{epi} \leftarrow [\ ]$ \\
    $\textbf{s} = \text{env.reset()}$ \\
    \While{$\textbf{s}$ is not a terminal state}{
      $
      \textbf{a} = \left\{
        \begin{array}{ll}
          \pi(\textbf{s}) + \textbf{noise} & \text{with probability } \mathcal{E}_R\\
          \pi(\textbf{s}) & \text{with probability } 1 - \mathcal{E}_R
        \end{array}
      \right.
      $ \\
      $\textbf{s}',r,done = \text{env.step}(\textbf{a})$ \\
      $\textbf{epi}$.append$([\textbf{s},\textbf{a},r,done,\textbf{s}'])$ \\
      $\textbf{s} \leftarrow \textbf{s}'$
    }
    $\textbf{episodes}$.append$(\textbf{epi})$ \\
    \If{$\textbf{epi}$ fails and with probability $\mathcal{E}_T$}{
      $\textbf{demo\_epi} \leftarrow [\ ]$ \\
      $\mathbf{c_s}, \mathbf{c_g} \leftarrow$ Start and goal configurations of the robot from $\textbf{epi}$ \\
      $\textbf{demo\_epi} \leftarrow \text{MP}(\mathbf{c_s}, \mathbf{c_g})$ \\
      $\textbf{episodes}$.append$(\textbf{demo\_epi})$
    }
  }
  \KwOut{$\textbf{episodes}$}
  \caption{\textbf{Generate Episodes with Mixed Exploration}}
  \label{alg:generate_episodes}
\end{algorithm}

Algorithm \ref{alg:generate_episodes} describes the process of generating episodes with mixed exploration. 
After generating $N$ episodes, $\mathcal{E}_R$ and $\mathcal{E}_T$ will be updated by 
\begin{align}
    \mathcal{E}_R = \text{max} &\left( \mathcal{E}_R - \Delta \mathcal{E}_R, 0 \right) \\
    \mathcal{E}_T = \text{min} &\left( \mathcal{E}_T + \Delta \mathcal{E}_T, 1 \right), 
\end{align}
where $\Delta \mathcal{E}_R > 0$ and $\Delta \mathcal{E}_T > 0$ are the changes applied to $\mathcal{E}_R$ and $\mathcal{E}_T$ respectively. 

\section{Experimental Setup}

A series of experiments are conducted in order to evaluate the performance of the proposed methods.
First the training and testing settings are defined followed by a comparison of different exploration strategies for RL training.
The learned motion planner from RL is also compared against classical motion planners.
Lastly, the policy is implemented and tested on a real robot to pass a 5.4mm suture needle, and all regrasps are done in a single pass. 

The needle selected for simulation is of radius $R= 5.4$mm.
To resemble the preferred grasping position and orientation of surgeons before throwing the suture needle, the goal is generated by setting $\alpha=\frac{4\pi}{3}$ and $d\in[6.5 \text{mm}, 8 \text{mm}]$, $\theta=0$, $\phi=\pi$ in the needle and grasping point frame respectively.
Meanwhile, the randomized initialization for the initial grasping is generated by sampling uniformly from $\alpha\in\left[\frac{11}{18}\pi, \frac{13}{18}\pi\right]$ and $d\in\left[6.5\text{mm}, 8\text{mm}\right]$, $\theta\in\left[0, 2\pi\right]$, $\phi\in\left[0, 0.4\pi\right]$ in the needle and grasping point frame respectively.
The regrasping end-effector is randomly initialized to be 13mm away from the center of the needle with a position variation $\sim\mathcal{U}(-1\ \text{mm}, 1\ \text{mm})$ and a orientation variation $\sim\mathcal{U}(-5^{\circ}, 5^{\circ})$ along some rotation axis. 
$\mathcal{U}(\cdot, \cdot)$ is a uniform distribution. 
Expert demonstrations for both the behavioral cloning and mixed exploration are generated using batch informed trees (BIT*)~\cite{gammell2015batch} due to its good performance in planning time.
The maximum horizon $T$ of the RL environment is 100, $\beta$ in the reward function is tuned to be $-0.02$, and $(\varepsilon_p, \varepsilon_q) = (1\text{mm}, 5^{\circ})$. 
The values of $(\varepsilon_p, \varepsilon_q)$ are empirically set to ensure a proper grasp. 
For the upcoming comparison studies, a test set of 300 randomized initialized grasps is generated and used for all the corresponding results.

The implementation of DDPG + BC is based on OpenAI Baselines~\cite{baselines}.
The actor and critic neural networks in DDPG are both multilayer perceptrons with 3 layers and 512 neurons per layer.
They are trained for a total of 500 epochs. 
Each epoch contains 10 iterations, where in each iteration, 5 episodes are collected in parallel to fill a replay buffer with a set size of $10^6$.
Per iteration, the actor and critic networks are updated a total of 200 times with a batch size of 256 and learning rate of $10^{-4}$ for both the actor and critic. 
Meanwhile, the target actor and critic networks are updated only once every 40 of those updates with a coefficient value of 0.95 for polyak-averaging.
The discount factor, $\gamma$, is set to 0.99, and an additional quadratic penalty for the actions is with a coefficient of 1.
The coefficient for the primary and cloning loss of the actor are both set to $10^{-3}$.
The total number of demonstration episodes generated for BC is 9900. 

\section{Results}
\subsection{Solving Needle Passing with Mixed Exploration}

To compare the proposed RL exploration algorithm with other methods, three separate policies are trained with the following settings in the needle passing environment: 
\begin{enumerate}
    \item DDPG + BC~\cite{nair2018overcoming} where: $\mathcal{E}^0_R = 0.3$, $\Delta \mathcal{E}_R = 6 \times 10^{-4}$
    \item DDPG + BC with targeted exploration~\cite{jurgenson2019harnessing} where: $\mathcal{E}^0_R = 0.02$, $\Delta \mathcal{E}_R = 0$, $\mathcal{E}^0_T=0.5$, $\Delta \mathcal{E}_T = 0$
    \item DDPG + BC with mixed exploration where: $\mathcal{E}^0_R = 0.3$, $\Delta \mathcal{E}_R = 6\times 10^{-4}$, $\mathcal{E}^0_T=0.1$, $\Delta \mathcal{E}_T = 4.5\times 10^{-3}$
\end{enumerate}
The DDPG + BC portion in all three methods uses the same network architecture and training hyper-parameters as previously listed.
Also, each setting is run three times with different random seeds. 

\begin{figure}[t]
    \centering
    \vspace{1.8mm}
    \includegraphics[trim=0cm 0cm 0cm 1.25cm, clip, width=0.48\textwidth]{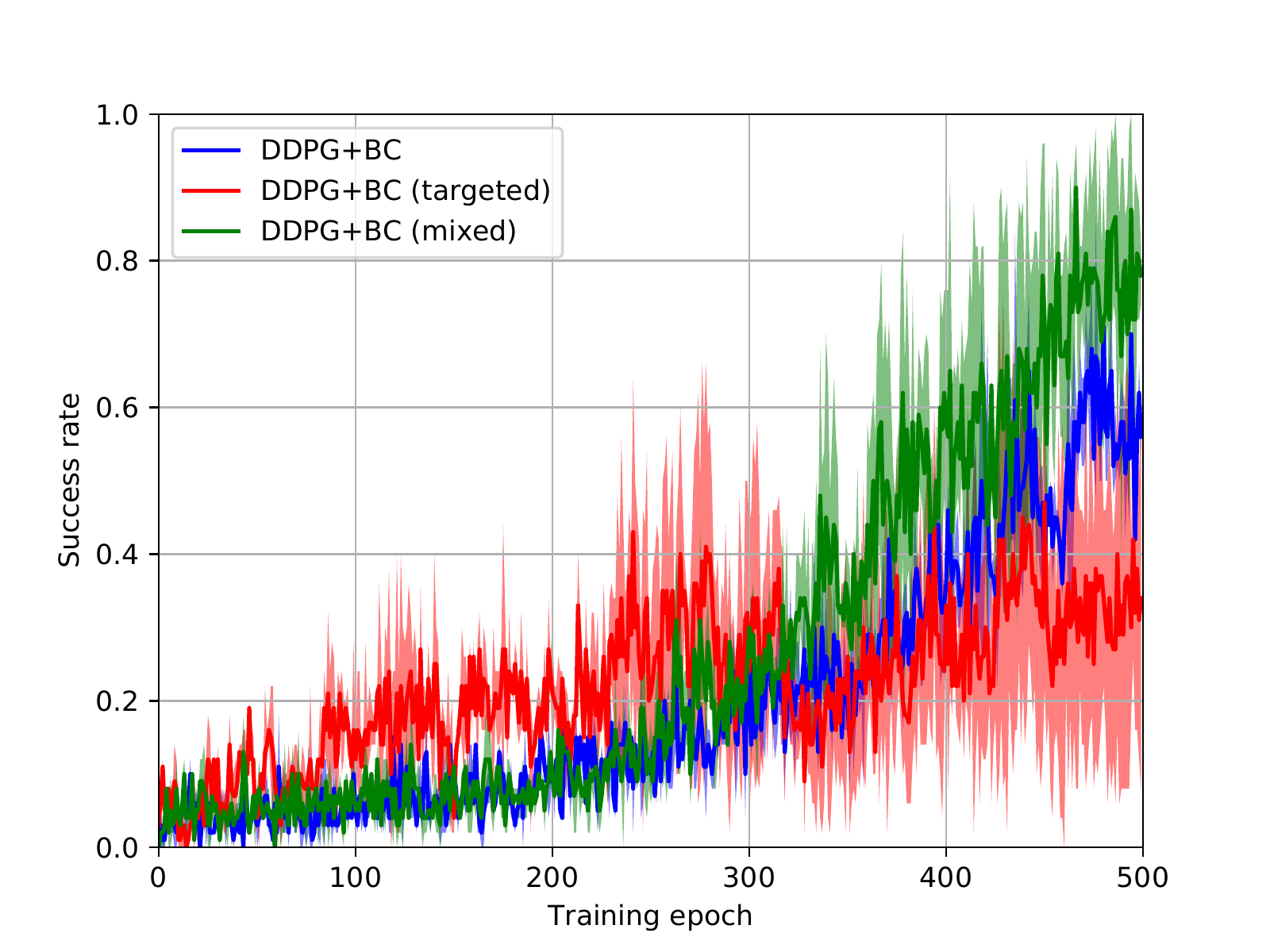}
    \caption{Training curves of different exploration strategies to solve the needle passing environment.}
    \label{fig:learning_curves_RL}
\end{figure}

\begin{table}[t]
\centering
\begin{tabular}{|c|c|c|c|}
    \hline
    \multirow{2}{*}{Algorithm} & \multirow{2}{*}{Success rate} & Planning time & Path length \\
     & & (s) & (mm) \\
    \hline\hline
    DDPG+BC & 0.71 & 0.0262 & \textbf{4.2} \\
    DDPG+BC (targeted) & 0.64 & 0.0233 & \textbf{4.2} \\
    DDPG+BC (mixed) & \textbf{0.97} & \textbf{0.0212} & \textbf{4.2} \\
    \hline
\end{tabular}
\caption{Success rate, average planning time and path length of different RL algorithms on the needle passing task}
\label{tab:performance_RL}
% \vspace{-0.1cm}
\end{table}

The training curves are shown in Fig. \ref{fig:learning_curves_RL}, and performance of the best trained policies are shown in Table \ref{tab:performance_RL}.
DDPG + BC (mixed) outperforms the other two methods with regards to success rate and planning time.
DDPG + BC (targeted) does reach a higher success rate at the beginning, but becomes unstable as the training goes on due to more heavily relying on imperfect experts.
Therefore, DDPG + BC is able to eventually surpass DDPG + BC (targeted) for success rate.

\subsection{Comparing Against Sampling-Based Motion Planners}
To show that DDPG + BC (mixed) can solve the suture needle passing task effectively and efficiently, a comparison against other sampling-based motion planning algorithms is conducted.
These algorithms include probabilistic roadmaps with the star strategy (PRM*)~\cite{kavraki1996probabilistic, karaman2011sampling}, rapidly exploring random trees with the star strategy (RRT*)~\cite{karaman2011sampling}, bidirectional fast marching trees (BFMT*)~\cite{starek2015asymptotically}, and BIT*~\cite{gammell2015batch}.
All of these algorithms are implemented on the needle passing environment using The Open Motion Planning Library \cite{sucan2012the-open-motion-planning-library}. These are top performing planners in OMPL.
Table \ref{tab:performance_all} summarizes their respective performances alongside the proposed learning based motion planning strategy in the needle passing environment.
The proposed method performs significantly better with regards to success rate and planning time.
The planners are also tested in reverse to highlight the speed up and improved success rate when doing so, hence showing why BIT* in reverse, which performed the best of the sampling-based planners, is used to generate demonstrations for the RL training.
%A more detailed view of BIT* and the proposed method is shown in Fig. \ref{fig:planning_time_path_length_bitstar_rl}.

\begin{table}[t]
\centering
\vspace{2mm}
\begin{tabular}{|c|c|c|c|}
    \hline
    \multirow{2}{*}{Algorithm} & \multirow{2}{*}{Success rate} & Planning time & Path length \\
     & & (s) & (mm) \\
    \hline\hline
    PRM* (F) & 0.66 & 103.342 & 5.2 \\
    PRM* (R) & 0.72 & 89.9856 & 5 \\
    RRT* (F) & 0.18 & 186.279 & 5.1 \\
    RRT* (R) & 0.2 & 147.6574 & 4.7 \\
    BFMT* (F) & 0 & -- & -- \\
    BFMT* (R) & 0 & -- & -- \\
    BIT* (F) & 0.66 & 3.8377 & 5.2 \\
    BIT* (R) & 0.72 & 2.5905 & 5 \\
    DDPG+BC (mixed) & \textbf{0.97} & \textbf{0.0212} & \textbf{4.2} \\
    \hline
\end{tabular}
\caption{Success rate, average planning time and path length of different motion planning algorithms on the needle passing task (F: forward planning, R: reverse planning)}
\label{tab:performance_all}
% \vspace{-1mm}
\end{table}

\begin{comment}
\begin{figure}[t]
    \centering
    \includegraphics[width=0.24\textwidth]{images/planning_time_bitstar_rl.pdf}
    \includegraphics[width=0.24\textwidth]{images/path_length_bitstar_rl.pdf}
    \caption{Planning time and path length of BIT* and RL policies (F: forward planning, R: reverse planning)}
    \label{fig:planning_time_path_length_bitstar_rl}
\end{figure}
\end{comment}

\subsection{Real World Experiment}
%\textcolor{red}{Add figure which shows how the needle reconstruction is converted to a feasible pose in the grasping gripper}

%The best trained RL policy for planning the regrasp is tested in the real world on a da Vinci Research Kit (dVRK) \cite{kazanzides2014open} with a suture needle of radius 5.4mm.
The best trained RL policy is tested in the real world on a da Vinci Research Kit (dVRK) \cite{kazanzides2014open} with a suture needle of radius 5.4mm.
The needle is initially grasped in one of the Patient Side Manipulators (PSM) arms from the dVRK using a Large Needle Driver (LND). 
The end-effector of this PSM arm is the grasping end-effector.
The end-effector of another PSM arm with an LND will act as the regrasping end-effector by following the RL policy to regrasp the needle.
In order to provide the state information for the RL policy, dVRK's stereo-endoscope is used which are 1080p and run at 30fps.
Both PSM arms are tracked from the stereo-endoscope using our previous work \cite{li2020super} which gives the pose of end-effectors in the camera frame.
Let $(\mathbf{p}^{c}_{e_g}, \mathbf{q}^c_{e_g})$ and $(\mathbf{p}^{c}_{e_r}, \mathbf{q}^c_{e_r})$ be the pose of the real-time tracked grasping and regrasping end-effectors respectively in the camera frame.
Then the ego-centric state for the regrasping end-effector can be computed as:
\begin{equation}
    \mathbf{H}^{e_g}_{e_r}(\mathbf{p}^{e_g}_{e_r},\mathbf{q}^{e_g}_{e_r}) = \mathbf{H}^{c}_{e_g}(\mathbf{p}^{c}_{e_g},\mathbf{q}^{c}_{e_g})^{-1} \mathbf{H}^{c}_{e_r}(\mathbf{p}^{c}_{e_r},\mathbf{q}^{c}_{e_r})
    \label{equ:convert_tool_tracking_to_ego_centric}
\end{equation}
In our previous work, we also showed effective control of the end-effector in the camera frame \cite{li2020super}.
Therefore, each action generated by the policy is converted to a target pose in the camera frame by solving $(\mathbf{p}^{c}_{e_r, t},\mathbf{q}^{c}_{e_r, t})$ using equations (\ref{equ:position_action}), (\ref{equ:quaternion_action}), and (\ref{equ:convert_tool_tracking_to_ego_centric}).
Then, following our previously developed controller, the end-effector of the regrasping arm in its own base frame is set to the pose $(\mathbf{p}^{b}_{e_r, t},\mathbf{q}^{b}_{e_r, t})$ via dVRK's built in inverse kinematic, which is computed by:
\begin{equation}
    \mathbf{H}^{b}_{e_r, t}(\mathbf{p}^{b}_{e_r, t},\mathbf{q}^{b}_{e_r, t}) = \mathbf{H}^{b}_{c}(\mathbf{p}^{b}_{c},\mathbf{q}^{b}_{c}) \mathbf{H}^{c}_{e_r, t}(\mathbf{p}^{c}_{e_r, t},\mathbf{q}^{c}_{e_r, t})
\end{equation}
where $\mathbf{H}^{b}_{c}(\mathbf{p}^{b}_{c},\mathbf{q}^{b}_{c})$ is updated in real-time by our previous surgical tool tracking method \cite{li2020super}.

\begin{figure}[t!]
\centering
\vspace{2mm}
    \begin{subfigure}{0.24\textwidth}
    \centering
        \includegraphics[height=2cm]{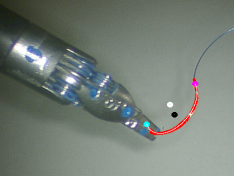}
        % \caption{Needle detection}
        \label{fig:needle_detection}
    \end{subfigure}%
    \begin{subfigure}{0.24\textwidth}
    \centering
        \includegraphics[height=2cm]{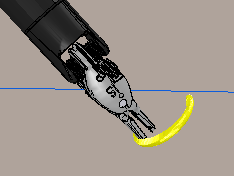}
        % \caption{Needle pose reconstruction}
        \label{fig:needle_reconstruction}
    \end{subfigure}
    \caption{Detected (left) and reconstructed (right) pose of a needle. Since the end-effector and needle are detected separately, their reconstructed poses might not form a feasible grasp.}
    \label{fig:needle_detection_reconstruction}
\end{figure}

%Similarly, the needle is reconstructed using the technique proposed by Lo et. al, which gives the pose of the needle in the camera frame~\cite{lo2002trip}. 
Similarly, the needle is reconstructed using the technique proposed in~\cite{lo2002trip}, which gives the pose of the needle in the camera frame.
Let $(\mathbf{p}^c_n, \mathbf{q}^c_n)$ be the reconstructed pose of the needle.
Combined with the previously described end-effector tracking, the reconstructed needle pose can be transformed to the grasping and regrasping end-effector frames by:
\begin{align}
    \mathbf{H}^{e_g}_{n}(\mathbf{p}^{e_g}_{n},\mathbf{q}^{e_g}_{n}) = \mathbf{H}^{c}_{e_g}(\mathbf{p}^{c}_{e_g},\mathbf{q}^{c}_{e_g})^{-1} \mathbf{H}^{c}_{n}(\mathbf{p}^{c}_{n},\mathbf{q}^{c}_{n})\\
    \mathbf{H}^{e_r}_{n}(\mathbf{p}^{e_r}_{n},\mathbf{q}^{e_r}_{n}) = \mathbf{H}^{c}_{e_r}(\mathbf{p}^{c}_{e_r},\mathbf{q}^{c}_{e_r})^{-1} \mathbf{H}^{c}_{n}(\mathbf{p}^{c}_{n},\mathbf{q}^{c}_{n})
\end{align}
hence giving the last components of the state information for the policy from the real world.
However, we experimentally found the reconstruction to be too inaccurate to directly apply as shown in Fig. \ref{fig:needle_detection_reconstruction}. This highlights that needle reconstruction is the weak point of transferring the RL policy to the real-world.
Therefore, to map the reconstructed pose $(\mathbf{p}^{e_g}_n, \mathbf{q}^{e_g}_n)$ to a valid initial grasp, this pose will be compared to a set of 1000 valid initial grasps $\{ (\mathbf{p}^{e_g}_n, \mathbf{q}^{e_g}_n)_1, (\mathbf{p}^{e_g}_n, \mathbf{q}^{e_g}_n)_2, \dots, (\mathbf{p}^{e_g}_n, \mathbf{q}^{e_g}_n)_{1000} \}$ collected from the simulated scene beforehand. 
The pose in this set that is closest to $(\mathbf{p}^{e_g}_n, \mathbf{q}^{e_g}_n)$ will be used for the policy. 
The distance is calculated in the same manner as equation (\ref{equ:pos_diff}) and (\ref{equ:quat_diff}). 

\begin{figure}[t!]
\centering
    \begin{subfigure}{0.24\textwidth}
    \centering
        \includegraphics[height=2cm]{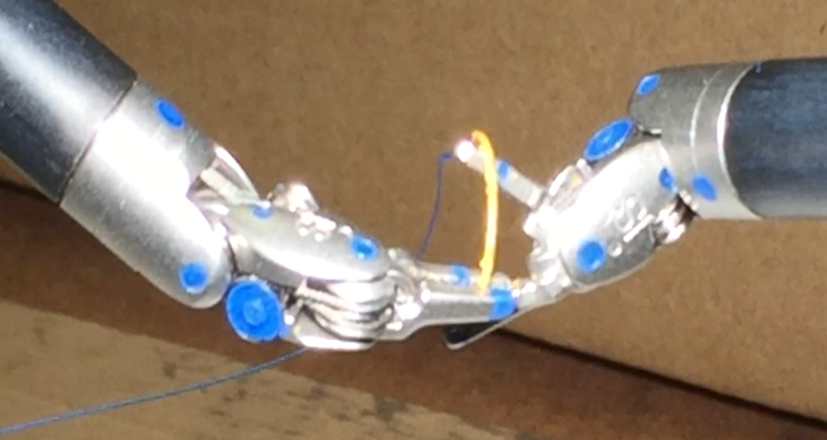}
        % \caption{Reach a joint limit}
        \label{fig:nreach_joint_limit}
    \end{subfigure}%
    \begin{subfigure}{0.24\textwidth}
    \centering
        \includegraphics[height=2cm]{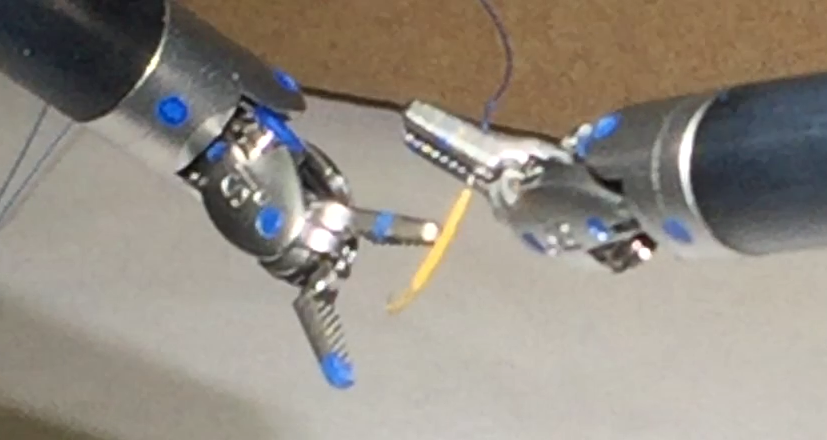}
        % \caption{Needle grasped too inside}
        \label{fig:grasped_too_inside}
    \end{subfigure}
    \caption{Failure cases of needle passing which are hitting a joint limit (left) and grasped too far inside (right)}
    \label{fig:needle_passing_failure}
\end{figure}

In the real-world experiments, the poses of an initial needle grasp and initial end-effectors are manually randomized.
The complete success rate for the needle passing is \textbf{73.3\%} from 15 trials, with an average planning time of 0.0846s and an average run time of 5.1454s.
Several example regrasps are shown in Fig. \ref{fig:cover_figure}.
In these trials, the main failures are caused by the needle reconstruction.
To test this, a secondary experiment is conducted where the initial grasp is preset to some known position and orientation.
During this experiment, the success rate of the needle passing is \textbf{90.5\%} from 21 trials, with an average planning time of 0.0807s and an average run time of 2.8801s.
In this case, the main failures come from the arm reaching a joint limit or needle regrasped too inside of the gripper, which are shown in Fig. \ref{fig:needle_passing_failure}. 

\section{Discussion and Conclusion}
In this work, we present a novel method for trajectory generation to conduct suture needle regrasping.
The task of regrasping is a critical and time consuming task during RAMIS. 
It is critical since the suture needle needs to be properly orientated and positioned to conduct an effective throw. Given the proposed work, one can combine it with the litany of work in automatic needle throwing to complete the autonomous suturing procedure.
%The proposed trajectory generation technique was validated on a dVRK with standard surgical tooling, hence making it a directly deployable technique.
%Furthermore, future researchers can investigate other extensions through the open-sourced suture needle environment on dVRL.
%For future work, we intend on extending this method into complete automation of multi-throw suturing.
Moving forward, the largest contributor to failed grasps is from inaccurate needle reconstruction.
This will be solved by applying Bayesian estimation under constraints \cite{simon2010kalman} where the constraint will be feasible grasping positions and orientations as described in this work.

\begin{comment}
\section*{Acknowledgement}
This research was supported by the Telemedicine and Advanced Technology Research Center (TATRC) T.R.O.N. program.
\end{comment}

\balance
\bibliographystyle{IEEEtran}
\bibliography{main}

\end{document}